\documentclass[10pt,twocolumn,letterpaper]{article}

\usepackage{fancyhdr}
\usepackage[textheight=1.618\textwidth,
lmargin=37mm, rmargin=22mm, driver=pdftex,
heightrounded, headsep=7mm,
footskip=11mm, vmarginratio=1:1]{geometry}

\usepackage{cvpr}
\usepackage{times}
\usepackage{epsfig}
\usepackage{graphicx}
\usepackage{amsmath}
\usepackage{amssymb}
\DeclareMathOperator*{\argmin}{argmin}

\newcommand*{\argminl}{\argmin\limits}

\def\fig#1{fig.\,\ref{#1}}

\usepackage{algorithm}
\usepackage{algorithmic}

\def\cvprPaperID{1341} 

\ifcvprfinal\fi
\begin{document}

\title{Parametric Image Segmentation of Humans\\ with Structural Shape Priors}

\author{Alin-Ionut Popa$^2$, Cristian Sminchisescu$^{1, 2}$\\
$^1$Department of Mathematics, Faculty of Engineering, Lund University\\ $^2$Institute of Mathematics of The Romanian Academy\\
{\tt\small alin.popa@imar.ro, cristian.sminchisescu@math.lth.se}
}

\maketitle


\lfoot{}
\rfoot{}
\cfoot{\thepage}



\begin{abstract}
The figure-ground segmentation of humans in images captured in natural environments is an outstanding open problem due to the presence of complex backgrounds, articulation, varying body proportions, partial views and viewpoint changes. In this work we propose class-specific segmentation models that leverage parametric max-flow image segmentation and a large dataset of human shapes. Our contributions are as follows: (1) formulation of a sub-modular energy model that combines class-specific structural constraints and data-driven shape priors, within a parametric max-flow optimization methodology that systematically computes all breakpoints of the model in polynomial time; 
(2) design of a data-driven class-specific fusion methodology, based on matching against a large training set of exemplar human shapes (100,000 in our experiments), that \emph{allows the shape prior to be constructed on-the-fly, for arbitrary viewpoints and partial views}. 
(3) demonstration of state of the art results, in two challenging datasets, H3D and MPII (where figure-ground segmentation annotations have been added by us), where we substantially improve on the first ranked hypothesis estimates of mid-level segmentation methods, by \emph{$20\%$, with hypothesis set sizes that are up to one order of magnitude smaller.}
\end{abstract}

\section{Introduction}

Detecting and segmenting people in real-world environments are central problems with applications in indexing, surveillance, 3D reconstruction and action recognition.  Prior work in 3D human pose reconstruction from monocular images\cite{urtasun2008sparse,ils_iccv11, ics-cvpr14}, as well as more recent, successful RGB-D sensing systems based on Kinect\cite{shotton11realtime} have shown that the availability of a figure-ground segmentation opens paths towards robust and scalable systems for human sensing. Despite substantial progress, the figure-ground segmentation in RGB images remains extremely challenging, because people are observed from a variety of viewpoints, have complex articulated skeletal structure, varying body proportions and clothing, and are often partially occluded by other people or objects in the scene. The complexity of the background further complicates matters, particularly as any limb decomposition of the human body leads to parts that are relatively regular but not sufficiently distinctive even when spatial connectivity constraints are enforced\cite{yang13}. Set aside appearance inhomogeneity and color variability due to clothing, which can overlap the background distribution significantly, it is well known that many of the generic, parallel line (ribbon) detectors designed to detect human limbs, fire at high false positive rates in the background. This has motivated work towards detecting more distinctive part configurations, without restrictive assumptions on part visibility (e.g. full or upper view of the person), for which poselets\cite{bourdev10} have been a successful example. However, besides relatively high false positive rates typical in detection, the transition from a bounding box of the person to a full segmentation of the human body is not straightforward. The challenge is to balance, on one hand, sufficient flexibility towards representing variability due to viewpoint, partial views and articulation, and, on the other hand, sufficient constraints in order to obtain segmentations that correspond to meaningful human shapes, all relying on region or structural human body part detectors that may only be partial or not always spatially accurate. 

In this work we attempt to connect two relevant, recent lines of work, for the segmentation of people in real images. We rely on bottom-up figure-ground generation methods and region-level person classifiers in order to identify promising hypothesis for further processing. In a second pass we set up informed constraints towards (human) class-specific figure-ground segmentation by leveraging skeletal information and data-driven shape priors computed on the fly by matching region candidates against exemplars of a large, recently introduced human motion capture dataset containing 3D and 2D semantic skeleton information of people, as well images and figure-ground masks from background subtraction (Human3.6M\cite{human36m}). By exploiting globally optimal parametric max-flow energy minimization solvers, this time based on a class dependent (as opposed to generic and regular) foreground seeding process \cite{gallo,kolmogorov_parametric_07,carreira_pami12}, we show that we can considerably improve the state of the art. To our knowledge this is one of the first formulations for class-specific segmentation that can handle multiple viewpoints and any partial view of the person, in principle. It is also one of the first to leverage a large dataset of human shapes, together with semantic structural information, which until recently, have not been available. We show that such constraints are critical for accuracy, robustness, and computational efficiency.

\subsection{Related Work}

The literature on segmentation is huge, even when considered only under sub-categories like top-down (class-specific) and bottom-up segmentation. Humans are of considerable interest to be devoted special methodology, if that proves to be effective\cite{LadickyTZ13,WangK11,GhiasiYRF14,XiaSFCY12,SigalB06,ferrari09,Andriluka2009,bourdev10,Zuffi:CVPR:2012,yang13,zuffiestimating13}. One approach is to consider shape as category-specific property and integrate it within models that are driven by bottom-up processing\cite{BoussaidK14,borenstein2002class,alpert2007image,objcut_pami09,Bray06Posecut,leibe08,levin2006learning,bourdev10,lempitsky2008image,pishchulin13cvpr,gavrila2013}. Pishchulin \textit{et al.} \cite{pishchulin13cvpr} develop pictorial structure formulations constrained by poselets, focusing on improving the response quality of an articulated part-based human model. The use of priors based on exemplars has also been explored, in a data-driven process. Both \cite{russell2009segmenting,rosenfeld2011extracting} focus on a matching process in order to identify exemplars that correspond to similar scene or object layouts, then used in a graph cut process that enforces spatial smoothness and provides a global solution. Our approach is related to such methods, but we use a novel data-driven prior construction, enforce structural constraints adapted to humans, and search the state space exhaustively by means of parametric max-flow. In contrast to priors used in \cite{russell2009segmenting,rosenfeld2011extracting}, which require a more repeatable scene layout, we focus on a prior generation process that can handle a diverse set of viewpoints and arbitrary partial views, not known a-priori, and different across the detected instances. 

Methods like \cite{kuettel2012figure} resemble ours in their reliance on a detection stage and the principle of matching that window representation against a training set where figure-ground segmentations are available, then optimizing an energy function based on graph-cuts. Our window representation contains additional detail and this makes it possible to match exemplars based on the semantic content identified. Our matching and shape prior construction are optimized for humans, in contrast to the generic ones used in \cite{kuettel2012figure} (which can however segment any object, not just people, as our focus here\footnote{Notice however that the methodology we propose would be applicable to other objects than people. Here we focus on people because only for them, for now, large training sets of segmented shapes with structural annotations are available, through Human3.6M\cite{human36m}. However as large datasets for other object categories emerge, we expect our methodology to generalize well. In this respect, our results on a challenging visual category, humans, are indicative of the performance bounds one can expect.}). We use large prior set of structurally annotated human shapes, and search the state space using a different, parametric multiple hypothesis scheme. Our prior construction uses, among other elements, a Procrustes alignment not unlike\cite{GuALYM12} but differently: (1) we use it for shape prior construction (input dependent, on the fly) within energy optimizer as opposed to object detection (classification, construction per class) in \cite{GuALYM12}, (2) we only use instances that align well with query reflecting accurate shape modeling, as opposed to fusing top-k instances to capture class variability in \cite{GuALYM12}. An alternative, interesting formulation for object segmentation with shape priors is branch-and-mincut\cite{Lempitsky_branch_08}, who propose a branch and bound procedure in the compound space of binary segmentations and hierarchically organized shapes. However, the bounding process used for efficient search in shape space would rely on knowledge of the type of shapes expected and their full visibility. We focus on a different optimization and modeling approach that can handle arbitrary occlusion patterns of shape. Our prior constraint for optimization is generated on the fly by fusing the visible exemplar components, following a structural alignment scheme.

Recently there has been a resurrection of bottom-up segmentation methods based on multiple proposal generation, with surprisingly good results considering the low-level processing involved. Some of these methods generate segment hypotheses either by combining the superpixels of a hierarchical clustering method\cite{arbelaez_pami,malisiewicz-bmvc07,van2011segmentation,Bro11b}, by varying the segmentation parameters\cite{endres_proposal_eccv2010} or by searching an energy model, parametrically, using graph cuts\cite{carreira_pami12,endres_proposal_eccv2010,kim2012shape,levinshtein10, MYP2011, DongCYY14}. Most of the latter techniques use mid-level shape priors for selection, either following hypothesis generation \cite{carreira_pami12,endres_proposal_eccv2010,levinshtein10} or during the process. Some methods provide a ranking, diversification and compression of hypotheses, using e.g. Maximal Marginal Relevance (MMR) diversification\cite{carreira_pami12,endres_proposal_eccv2010}, whereas others report an unordered set\cite{kim2012shape,levinshtein10}. Hypothesis pool sizes in the order of 1,000-10,000 range in the expansionary phase, and compressed models of 100-1,000 hypotheses following the application of trained rankers (operating on mid-level features extracted from segments) with diversification, are typical, with variance due to image complexity and edge structure. While prior work has shown that such hypotheses pools can contain remarkably good quality segments ($60-80\%$ intersection over union, IoU, scores are not uncommon) this leaves sufficient space for improvement particularly since sooner or later, one is inevitably facing the burden of decision making: \emph{selecting one hypothesis to report}. It is then not uncommon for performance to sharply drop to $40\%$. This indicates that constraints and prior selection methods towards more compact, better quality hypothesis sets are necessary. Such issues are confronted in the current work.

\section{Methodology}

We will consider an image as $I:{\cal V}\rightarrow R^3$ , where ${\cal V}$ represents the set of nodes, each associated with a pixel in the image, and the range is the associated intensity (RGB) vector. The image is modeled as a graph $G = ({\cal V},{\cal E})$. We partition the set of nodes in ${\cal V}$ into two disjoint sets of ${\cal V}_f$ and ${\cal V}_b$ which represent the assignments of pixels to foreground and background, respectively. ${\cal E}$ is the subset of edges of the graph $G$ which reflects the connections between adjacent pixels. The formulation we propose will rely on object (or foreground) structural skeleton constraints obtained from person detection and 2D localization (in particular the identification of keypoints associated with the joints of the human body, and the resulting set of nodes corresponding to the human skeleton, obtained by connecting keypoints, $T \subseteq {\cal V}$), as well as a data-driven, human shape fusion prior $S:{\cal V} \rightarrow [0,1]$, constructed ad-hoc by fusing similar configurations with the one detected, based on a large dataset of human shapes with associated 2D skeleton semantics (see our \S\ref{sec:psp} for details). The energy function defined over the graph $G$, $X=\cup \{x_u\}$ is:

\begin{align}
\label{eqn:energyFunction}
E_{\lambda}(X) = \sum_{u \in {\cal V}} U_{\lambda}(x_u) + \sum_{(u,v) \in {\cal E}} V_{uv}(x_u,x_v)
\end{align}
where 
\begin{eqnarray*}\label{eqn:unary}
U_{\lambda}(x_u) &=& D_{\lambda}(x_u) + S(x_u)\\
\end{eqnarray*}
\noindent with $\lambda \in \mathbb{R}$, and unary potentials given by semantic foreground constraints ${\cal V}_f \leftarrow T$:
\begin{equation}
D_{\lambda}(x_u) = \left\{
    \begin{array}{l l}
                                  0 & \quad \mbox{if $x_u=1$, $u \notin {\cal V}_b$} \\
                                    \infty & \quad \mbox{if $x_u=1$, $u \in {\cal V}_b$} \\
                                    \infty & \quad \mbox{if $x_u=0$, $u \in {\cal V}_f$} \\
                                    f(x_u) + \lambda & \quad \mbox{if $x_u=0$, $u \notin {\cal V}_f$} \\
\end{array} \right.
\end{equation}

The foreground bias is implemented as a cost incurred by the assignment of non-seed pixels to background, and consists of a pixel-dependent value $f(x_u)$ and an uniform offset $\lambda$. Two different functions $f(x_u)$ are used alternatively. The first is constant and equal to $0$, resulting in a uniform (variable) foreground bias. The second function uses color. Specifically, RGB color distributions
$p_f(x_u)$ on seed ${\cal V}_f$ and $p_b(x_u)$ on seed ${\cal V}_b$ are estimated and derive $f(x_u) =
\ln \frac{p_f(x_u)}{p_b(x_u)}$. 
The probability distribution of pixel $j$ belonging to the foreground is defined as $p_f(i)=\exp(-\gamma \cdot \min_j(||I(i) - I(j)||))$, with $\gamma$ a scaling factor, 
and $j$ indexes representative pixels in the seed region, selected as centers resulting from a \emph{k}-means algorithm ($k$ is set to $5$ in all of our experiments). The background probability is defined similarly.

The pairwise term $V_{uv}$ penalizes the assignment of different labels to similar neighboring pixels:
\begin{equation}
\label{pairwise_eq}
V_{uv}(x_u, x_v) = \left\{
    \begin{array}{l l}
    0 & \quad \mbox{if $x_u = x_v$} \\
    g(u,v) & \quad \mbox{if $x_u \ne x_v$} \\
    \end{array} \right.
\end{equation}
with similarity between adjacent pixels given by $g(u, v) = \exp \Bigl[ -
\frac{\max(Gb(u), Gb(v))}{\sigma^2} \Bigr] $.  $Gb$ returns the output of
the multi-cue contour detector \cite{globalPB08,leordeanu_eccv12} at a pixel. The \textit{boundary sharpness} parameter $\sigma$ controls the smoothness of the pairwise term.

The energy function defined by \eqref{eqn:energyFunction} is submodular and can be optimized using parametric max-flow, in order to obtain all breakpoints of $E_{\lambda}(X)$ as a function of $(\lambda,X)$ in polynomial time.

\begin{figure*}[!htb]
\begin{center}
\includegraphics[width=\textwidth]{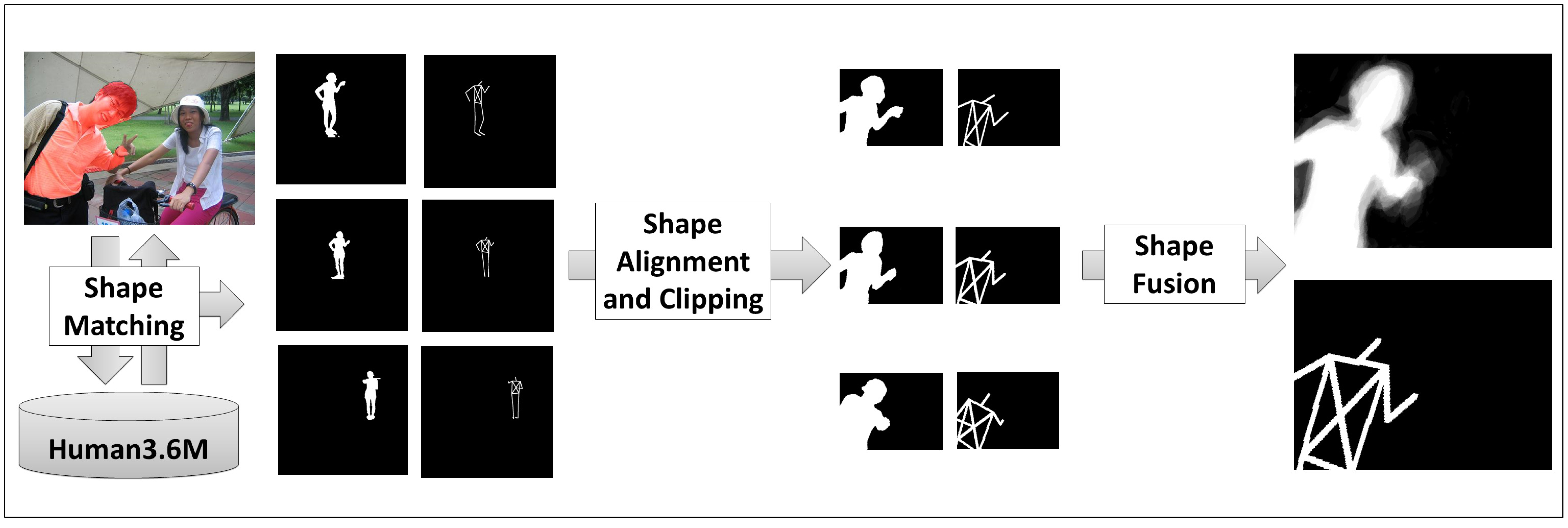}\\
\includegraphics[width=\linewidth]{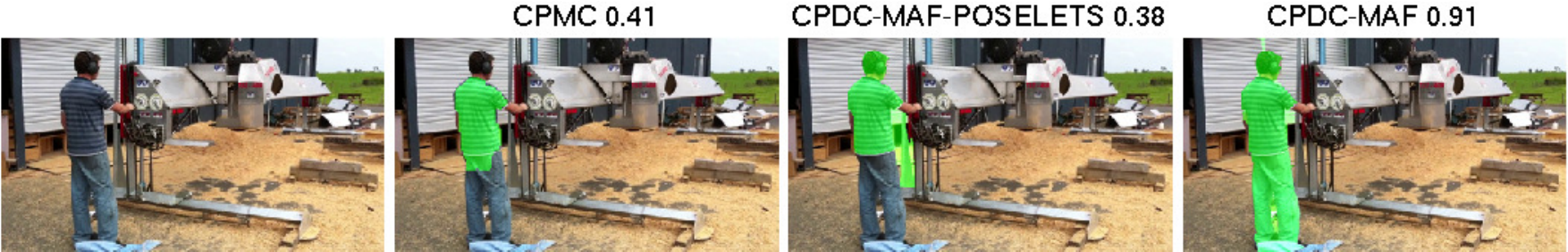}\\
\includegraphics[width=\linewidth]{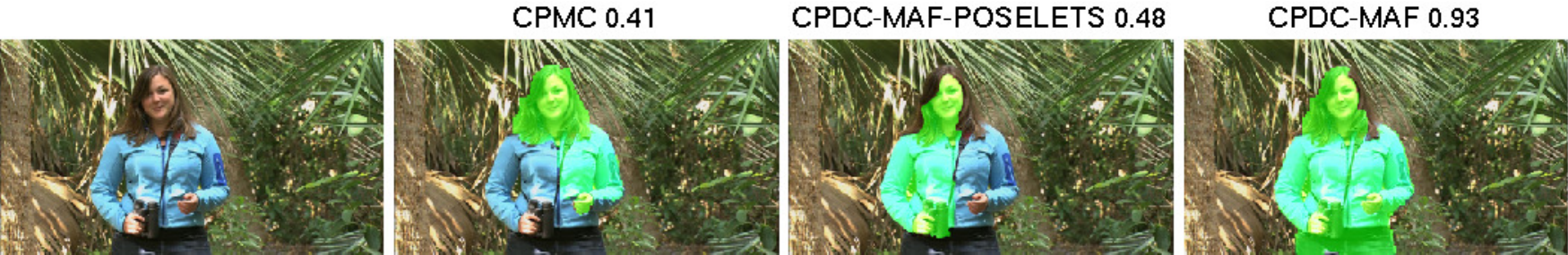}
\end{center}
\vspace{-3mm}
\caption{{\it First row:} Our Shape Matching Alignment Fusion (MAF) construction based on semantic matching, structural alignment and clipping, followed by fusion, to reflect the partial view. Notice that the prior construction allows us to match partial views of a putative human detected segment to fully visible exemplars in Human3.6M. This allows us to handle arbitrary patterns of occlusion. We can thus create a well adapted prior, on the fly, given a candidate segment. {\it Second and third rows:} Examples of segmentations obtained by several methods (including the proposed ones), with intersection over union (IoU) scores and ground truth shown. See \fig{fig:cpdcBestResult} for additional image segmentation results.}
\label{fig:cpdpFlow}
\end{figure*}

\begin{figure}[!htb]
\begin{center}
\includegraphics[width=0.9\linewidth]{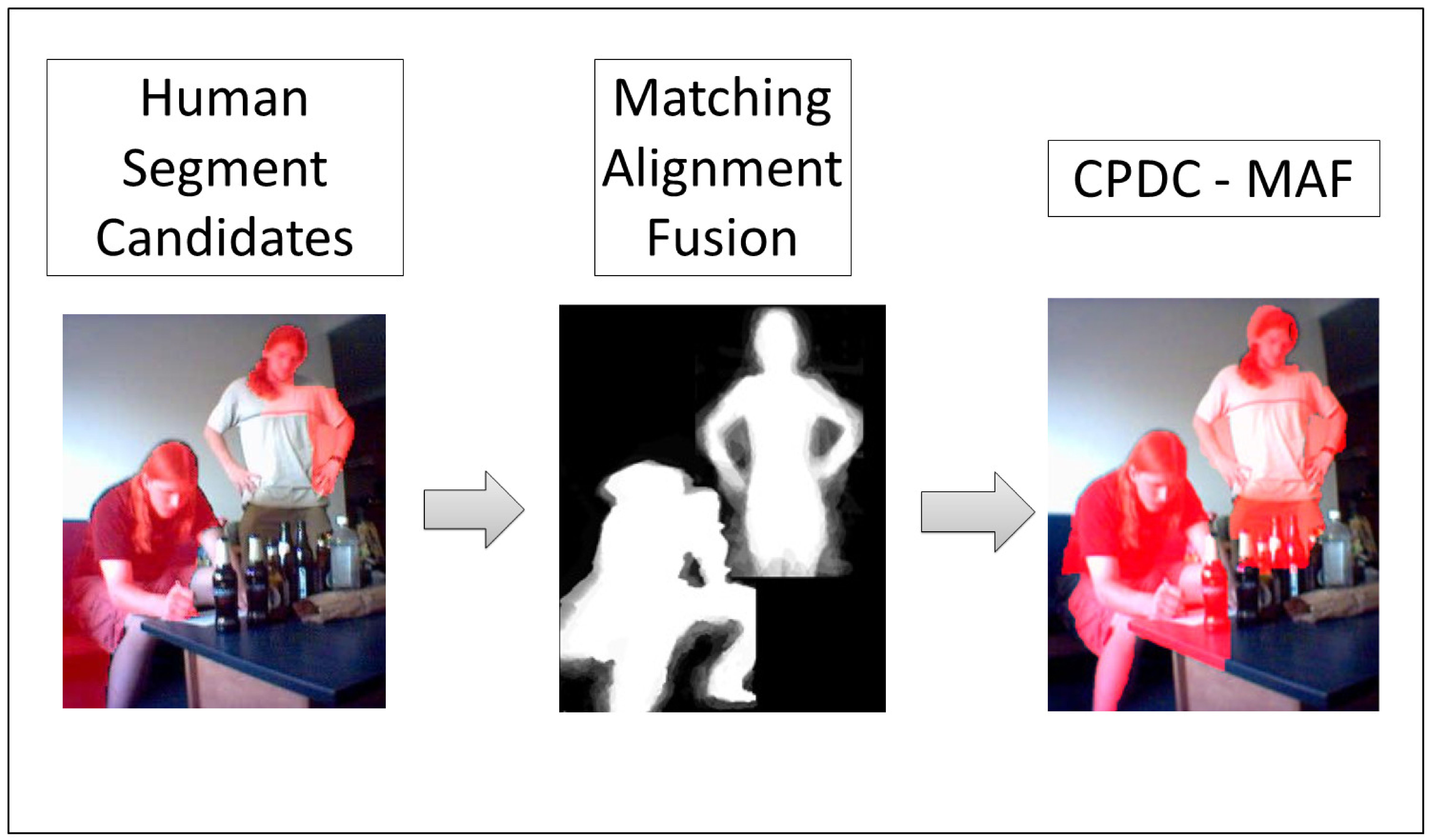}
\end{center}
\vspace{-3mm}
\caption{Processing steps of our segmentation methods based on Constrained Parametric Problem Dependent Cuts (CPDC) with Shape Matching, Alignment and Fusion (MAF). }
\label{fig:pipelineFlow}
\end{figure}

Given the general formulation in \eqref{eqn:energyFunction} and \eqref{eqn:unary}, the key problems to address are: {\bf (a)} the identification of a putative set of person regions and structural constraints hypotheses $T$; {\bf (b)} the construction of an effective, yet flexible data-driven human shape prior $S$, based on a sufficiently diverse dataset of people shapes and skeletal structure, given estimates for $T$. {\bf (c)} minimization of the resulting energy model \eqref{eqn:energyFunction}. We address (a) without loss of generality, using a human region classifier (any other set of structural, problem dependent detectors can be used, here e.g. face and hand detectors based on skin color models or poselets). 
We address (b) using methodology that combines a large dataset of human pose shapes and body skeletons, collected from Human3.6M\cite{human36m} with shape matching, alignment and fusion analysis, in order to construct the prior on the fly, for the instance being analyzed. We refer to a model that leverages both problem-dependent structural constraints $T$ and a data-driven shape prior $S$, in a \emph{single joint optimization problem}, as \emph{Constraint Parametric Problem Dependent Cuts with Shape Matching. Alignment and Fusion (CPDC-MAF)}. The integration of bottom-up region detection constraints with a shape prior construction is described in \S\ref{sec:psp}. The CPDC-MAF model can be optimized in polynomial time using parametric max-flow, in order to obtain all breakpoints of the associated energy model (addressing c). 

\subsection{Data-Driven Shape Matching, Alignment and Fusion (MAF)}\label{sec:psp}
\label{subseq:ddphsp}

We aim to obtain an improved figure-ground segmentation for persons by combining bottom-up and top-down, class specific information. We initialize our proposal set using CPMC\cite{carreira_pami12}. While any figure-ground segmentation proposal method can be employed, in principle, we chose CPMC due to its performance and because our method can be viewed as a generalization with problem dependent seeds and shape priors. We filter the top $N$ segment candidates using an O2P\cite{carreira_eccv12}-region classifier trained to respond to humans, using examples from Human3.6M, to obtain ${\cal D}=\{d_i=\{\mathbf{z}, \mathbf{b}\}, | i=1,\ldots N\}$. Each candidate segment is represented by a binary mask $\mathbf{z}_{i}$, $1$ stands for foreground and $0$ stands for background and a bounding box $\mathbf{b} \in \mathbb{R}^4$ where $\mathbf{b} = (m, n, w, h)$. $m$ and $n$ represent the image coordinates of the bottom left corner of the bounding box, $w$ and $h$ represents its width and its height. 

We will use the set of human region candidates in order to match against a set of human shape and construct a shape prior.
There are challenges however, particularly being able to: {\bf (1)} access a sufficiently representative set of human shapes to construct the prior, {\bf (2)} be sufficiently flexible so that human shapes from the dataset, which are very different from the shape being analyzed, would not negatively impact estimates, {\bf (3)} handle partial views---while we rely on bottom-up proposals that can handle partial views, the use, in contrast, of a shape prior that can only represent, e.g. full or upper-body views, would not be effective.

We address: (1) by employing a dataset of 100,000 human shapes together with the corresponding skeleton structure, sub-sampled from the recently created Human3.6M dataset\cite{human36m}; (2) by employing a matching, alignment and fusion technique between the current segment and the individual exemplar shapes in the dataset. Shapes and structures which cannot be matched and aligned properly are discarded; (3) by leveraging the implicit correspondences available across training shapes, at the level of local shape matches, by only aligning and warping those components of the exemplar shapes that can be matched to the query, at the level of joints.
A sample flow of our entire method can be visualized in figure \ref{fig:cpdpFlow} first row and figure \ref{fig:pipelineFlow}.

\vspace{3mm}
\noindent{\bf Boundary Point Sampling:} Given a bottom-up figure-ground proposal represented as a binary mask $\mathbf{z} \in \cal{D}$, we sample through the image coordinates of the boundary points of the foreground segment. Thus we obtain a set of 2D points $\mathbf{p}_{j}, j=1,\ldots,K$ with $\mathbf{p}_{j} \in \mathbb{R}^2$ where  $\mathbf{p}_{j} = (x_j, y_j)$. We loop through the shapes of our human shape dataset Human3.6M and for each shape we rotate and scale it so that it has the same orientation and scale as the foreground candidate segment and sample through it boundary points. Thus we obtain a set of 2D points $\mathbf{q}_{jl}, j=1,\ldots,K$, with $l=1,\ldots,L$, where $L$  represents the number of poses in the shape-pose dataset, in our case $L=100,000$. 

\vspace{3mm}

\noindent{\bf Shape Matching and Transform Matrix:} 
We employ the shape context descriptor\cite{BelongieMP02} at each position $\mathbf{p}_{j}$ from the candidate segment and each position $\mathbf{q}_{jl}$ from each shape from the dataset. We evaluate a $\chi^2$ distance on the resulting descriptors to select the indexes $l$ with enough well-matched of boundary points such that we could estimate an affine transform. 

We apply a 2D Procrustes transform with 5 degrees of freedom (rotation, anisotropic scaling including reflections, and translation) on $\mathbf{q}_l$ in order to align each shape in the dataset with the corresponding boundary points. This will result in a 3x3 transformation matrix $\mathbf{W}_l$ and an error for the transform $e_l$ which represents the Euclidean distance between the boundary points $\mathbf{p}_{j}$ and the Procrustes transformed ones, $\mathbf{W}_l \cdot \mathbf{q}_{lj}$, in the image plane. 

\vspace{3mm}

\noindent{\bf Prior Shape Selection and Warping:} In order to determine which prior shapes are relevant for the current detected query, we identified the subset of indexes in the dataset ${\cal T}$ which correspond to transformation errors that are smaller than a given threshold $\epsilon$. Thus, we obtain the corresponding figure-ground masks $\mathbf{m}_t, t \in {\cal T}$. For each mask $\mathbf{m}_t$ we selected the coordinates of foreground pixels and warp them using the transform matrix computed using the 2D joint coordinates transformation. We apply the same procedure to the attached skeleton configuration of the corresponding mask. Thus, we obtain the coordinates of the foreground pixels for the transformed mask, $\Phi_t$ and the transformed skelet coordinates $\Psi_t$.

\vspace{3mm}

\noindent{\bf Prior Shape Fusion:} We compute the mean of the entire set of transformed masks, $\Phi_t$, thus obtaining a MAF prior, $S$ corresponding to the detection $d$ as seen in figure \ref{fig:cpdpFlow}, second row. The values of the shape prior mask range from $0$ to $1$, background and foreground probabilities, respectively. Also we compute the mean of the entire set of transformed skeletons $\Psi_t$, thus obtaining a configuration of keypoints $\mathbf{B} \in \mathbb{R}^{3\times15}$ with $\mathbf{B}_j = (x, y, 1)$ where $x$ and $y$ represent the image coordinates of the  warped joint from Human3.6M. This could be used to obtain problem dependent mask $\mathbf{m}$ as follows. Initially we set the mask to have the same dimension as the entire image, filled with $0$. We use Bresenham's algorithm to draw a line between the semantically adjacent joints, for example: left elbow - left wrist, right hip - right knee, and so on. We assign the set of skeleton nodes to the foreground as $T=\{i \in {\cal V}| \mathbf{m}(i)=1\}$. This entire procedure of obtaining the shape prior information (mask and skeleton) is illustrated in algorithm \ref{alg:ShapePriorAlgorithm}.

\begin{algorithm}
\caption{Calculate $S$ and ${\mathbf{B}}$ (Shape Matching, Alignment and Fusion, {\bf MAF})}
\begin{algorithmic} 
\REQUIRE  
\STATE $d_i=\{\mathbf{z}, \mathbf{b}\}$
\STATE $\mathbf{d}_l, l=1,\ldots,L$ - 2D joint positions (Human3.6M)
\STATE $\mathbf{m}_l, l=1,\ldots,L$ - figure-ground masks (Human3.6M)
\STATE $L$ - number of poses (Human3.6M, use $L=100,000$)
\STATE $\epsilon$ - threshold value for transform error
\STATE $\mathbf{f}(\cdot)$ - shape context descriptor
\STATE $\mu$ - threshold value for $\chi^2$ for shape context descriptors
\ENSURE {$S$, ${\mathbf{B}}$}
\STATE Sample boundary points $\mathbf{p}_{j}, j=1,\ldots,K$ on $\mathbf{z}$
\FOR{$l \in {\cal L}$} 
\STATE Sample $K$ boundary points $\mathbf{q}_{jl}, j=1,\ldots,K$ on $\mathbf{m}_l$
\STATE $J=\{ (x,y) \in \mathbb{N}^2 | \chi^2(\mathbf{f}(\mathbf{q}_{xl}), \mathbf{f}(\mathbf{p}_{y})) < \mu\}$
\IF{$|J| > 2$}
\STATE {$\mathbf{a}_{jl}(\mathbf{W})=\mathbf{p}_{j} - \mathbf{W} \cdot \mathbf{q}_{jl}$}
\STATE {$\mathbf{W}_l = \argminl_{\mathbf{W}}\frac{1}{|K|}\sum_{\substack{j \in K}} \mathbf{a}_{jl}(\mathbf{W})^\top \mathbf{a}_{jl}(\mathbf{W})$}
\STATE {$e_l = \frac{1}{|K|}\sum_{\substack{j \in K}}\mathbf{a}_{jl}(\mathbf{W}_l)^\top \mathbf{a}_{jl}(\mathbf{W}_l)$}
\ELSE 
\STATE {$e_l=\infty$}
\ENDIF
\ENDFOR
\STATE {${\cal T} = \{l \in {\cal L} | e_l < \epsilon\} $}
\FOR{$t \in {\cal T}$} 
\STATE {${\cal V}_f$ - foreground pixels of $\mathbf{m}_t$}, {${\cal V}_b$ - background pixels of $\mathbf{m}_t$}, {${\cal V} = {\cal V}_b \cup {\cal V}_f$}\\
\FOR {$\mathbf{u} \in {\cal V}$}
\IF {$\mathbf{u} \in {\cal V}_f$} 
\STATE{$\Phi_t(\mathbf{W}_t \cdot \mathbf{u}) = 1$}  
\ELSE 
\STATE{$\Phi_t(\mathbf{W}_t \cdot \mathbf{u}) = 0$} 
\ENDIF
\ENDFOR
\STATE {$\Psi_t=\mathbf{W}_t \cdot \mathbf{d}_l$}
\ENDFOR
\STATE {$S = \frac{1}{|{\cal T}|}\sum_{\substack{t \in {\cal T}}}\Phi_t$}
\STATE {$\mathbf{B} = \frac{1}{|{\cal T}|}\sum_{\substack{t \in {\cal T}}}\Psi_t$}
\end{algorithmic}
\label{alg:ShapePriorAlgorithm}
\end{algorithm}

\section{Experiments}

We test our methodology on two challenging datasets: H3D\cite{PoseletsICCV09} which contains 107 images and MPII \cite{andriluka14cvpr} with 3799 images. In all cases we have figure-ground segmentation annotations available. For the MPII dataset, we generated figure-ground human segment annotations ourselves.
Both the H3D and the MPII datasets contain both full and partial views of persons and self-occlusion and are extremely challenging.

We run several segmentation algorithms including CPMC\cite{carreira_pami12} as well as our proposed CPDC-MAF where we use bottom-up person region detectors trained on Human3.6M and using region descriptors based on O2P\cite{carreira_eccv12}. We also constructed a model referred to as CPDC-MAF-POSELETS, built using problem dependent seeds based on a 2D pose detector instead of proposed segments from a figure-ground segmentation algorithm. While any methodology that provides body keypoints (parts or articulations) is applicable, we chose the poselet detector because it provides results under partial views of the body, or self occlusions of certain joints together with joint position estimates. Conditioned on a detection, we apply the same idea as in our CPDC-MAF, except that we use the detected skeletal keypoints to match against the exemplars in the Human3.6M dataset. A matching process based on semantic keywords (the body joints) is explicit, immediate (since joints are available both for the putative poselet detector and for the exemplar shapes in Human3.6M) and arguably simpler than matching shapes in the absence of skeletal information. The downside is that when the poselet detection is incorrect, the matching will also be (notice that alignments with high score following matching are nevertheless discarded within the MAF process). 

For CPDC-MAF, we initialize, bottom-up, by using candidate segments from CPMC pool, selected based on their $\mathbf{person}$ ranking score after applying the O2P classifier. This is followed by a non-maximum suppression step were we remove the pair of segments with an overlap above $0.25$. We use the MAF process to reject irrelevant candidates and to build shape prior masks and skeleton configuration seeds for the segments with good matching produced by shape context descriptors. On each resulting shape prior and skeleton seeds we run the CPDC-MAF model with the resulting pools from each candidate segment merged to obtain the human region proposals for an entire image. 

For each testing setup, we report the mean values (computed over the entire testing dataset) of the intersection over union (IoU) scores for the first segment in the ranked pool and the ground-truth figure-ground segmentation for each image. We also report the mean values of the IoU scores for the pool segment with the best IoU score with the ground-truth figure ground segmentation. 

\begin{figure*}[!htb]
\begin{center}
    \includegraphics[width =0.95\textwidth]
{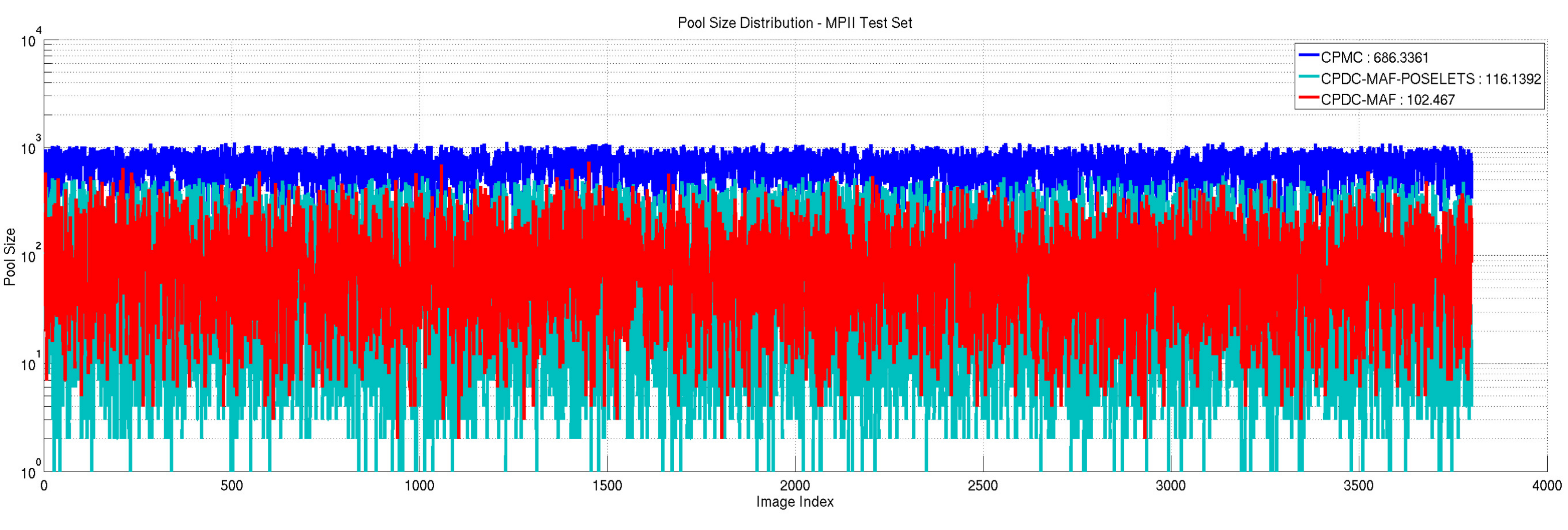} ~
\end{center}
   \caption{Dimension of segmentation pool for MPII and various methods along with average pool size (in legend). Notice significant difference between the pool size values of CPDC-MAF-POSELETS and CPDC-MAF compared to the ones of CPMC. CPMC pool size values maintain an average of 700 units, whereas the pool sizes of CPDC-MAF and CPDC-MAF-POSELETS are considerably smaller, around 100 units.}\label{fig:poolMPI}

\end{figure*}

\begin{figure*}[!htb]
\begin{center}
    \includegraphics[width =0.95\textwidth]
{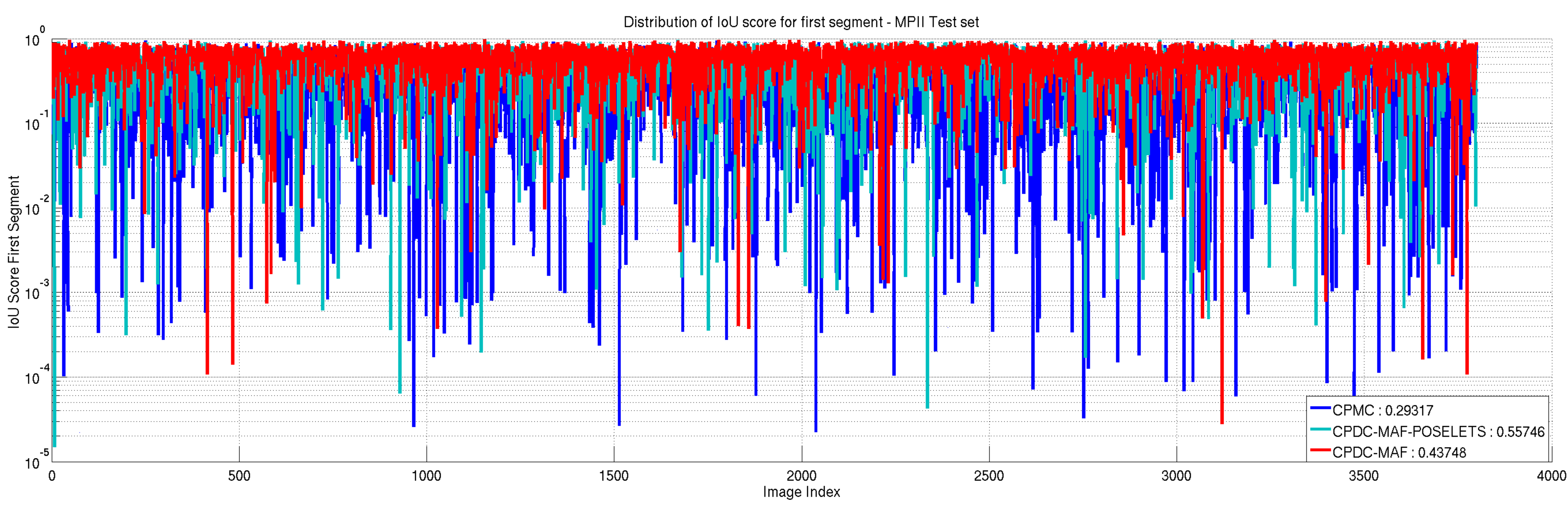} ~
\end{center}
   \caption{IoU for the first segment from the ranked pool in MPII. 
The values for CPMC and CPDC-MAF-POSELETS have higher variance compared to CPDC-MAF resulting in the performance drop illustrated by their average.}\label{fig:firstMPI}

\end{figure*}

\begin{figure*}[!htb]
\begin{center}
\scalebox{1}{
	\begin{tabular}{c}
   {\includegraphics[width=\linewidth]{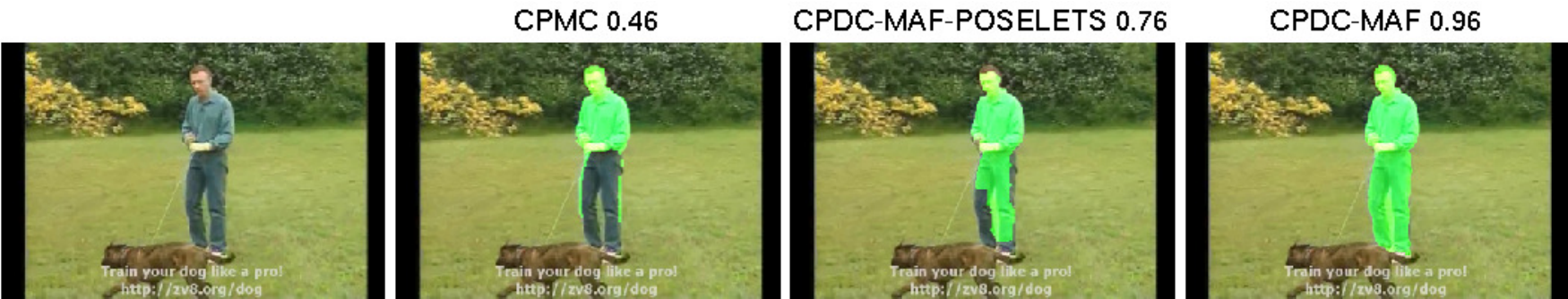}} \\
   {\includegraphics[width=\linewidth]{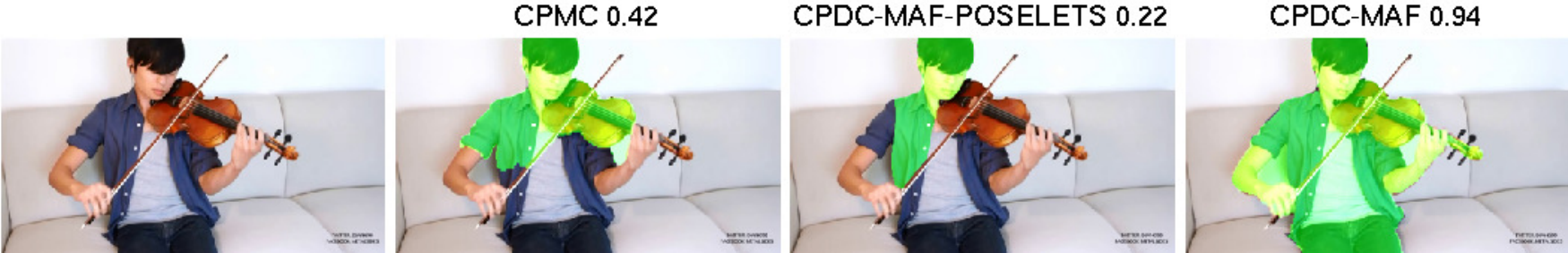}} \\
   {\includegraphics[width=\linewidth]{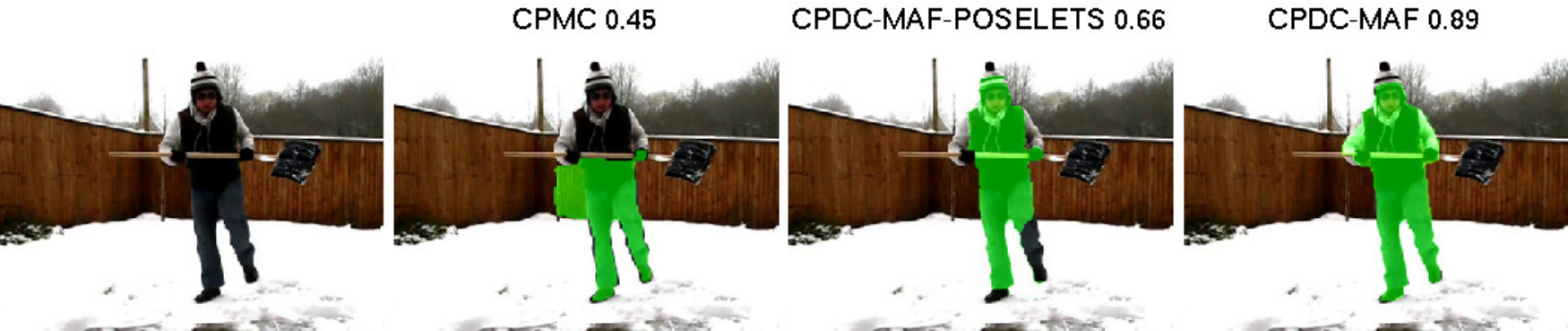}} \\
   {\includegraphics[width=\linewidth]{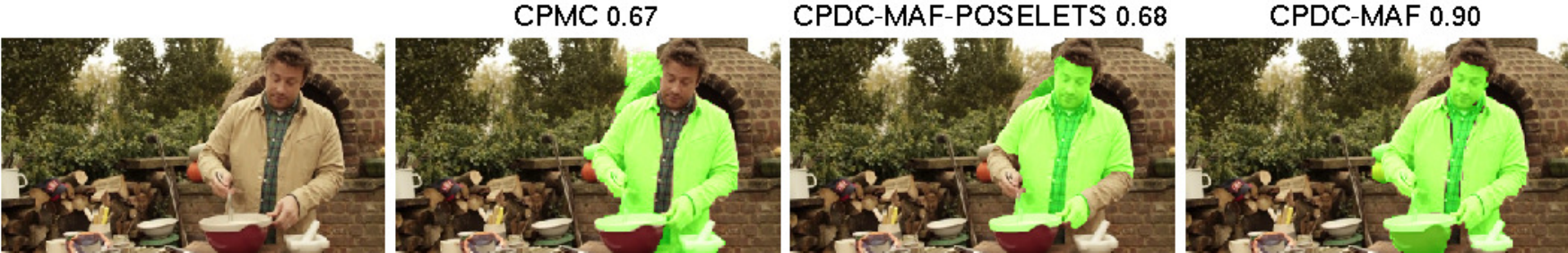}} \\
   {\includegraphics[width=\linewidth]{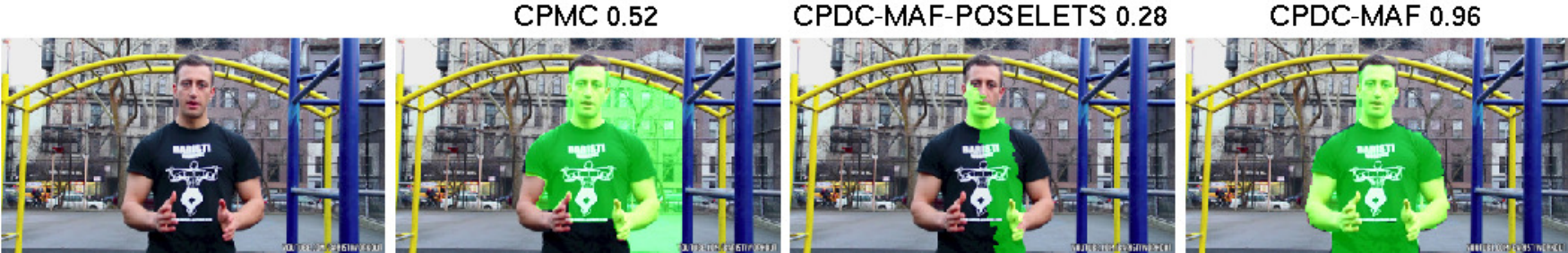}}
	\end{tabular}   
   }
\end{center}
\caption{Segmentation examples for various methods. From left to right, original image, CPMC with default settings on person's bounding box, CPDC-MAF-POSELET and CPDC-MAF. See also tables \ref{tab:tableResults} for quantitative results. {\it Please check our supplementary material for additional image results}.}
\label{fig:cpdcBestResult}
\end{figure*}

Results for different datasets can be visualized in table \ref{tab:tableResults}. In turn, figures \ref{fig:poolMPI}, \ref{fig:firstMPI} show plots for the size of the segment pools and IoU scores for highest ranked segments generated by different methods, with image indexes sorted according to the best performing method (CPDP-MAF). Qualitative segmentation results for the various methods tested are given in figure \ref{fig:cpdcBestResult}. 

\begin{table}[htb]
\begin{center}
\begin{tabular}{|c||c|c|c|}
\hline
Method & \multicolumn{3}{c|}{H3D Test Set\cite{PoseletsICCV09}} \\
\hline
& First & Best & Pool size \\
\hline
CPMC\cite{carreira_pami12} & 0.54 & 0.72 & 783 \\
\hline
CPDC - MAF & 0.60 & 0.72 & 77 \\
\hline
CPDC - MAF - POSELETS & 0.53 & 0.6 & 98 \\
\hline
\hline
& \multicolumn{3}{c|}{MPII Test Set\cite{andriluka14cvpr}}\\
\hline
& First & Best & Pool size \\
\hline
CPMC\cite{carreira_pami12} & 0.29 & 0.73 & 686 \\
\hline
CPDC - MAF & 0.55 & 0.71 & 102 \\
\hline
CPDC - MAF - POSELETS & 0.43 & 0.58 & 114 \\
\hline
\end{tabular}
\end{center}
\caption{Accuracy and pool size statistics for different methods, on data from H3D and MPII. We report average IoU over test set for the first segment of the ranked pool and the ground-truth figure-ground segmentation (\textit{First}), the average IoU over test set of the segment with the highest IoU with the ground-truth figure-ground segmentation (\textit{Best}) and average pool size (\textit{Pool Size}).}
\label{tab:tableResults}
\end{table}

\section{Conclusions}

We have presented class-specific image segmentation models that leverage human body part detectors based on bottom-up figure-ground proposals, parametric max-flow solvers, and a large dataset of human shapes. Our formulation leads to a sub-modular energy model that combines class-specific structural constraints and data-driven shape priors, within a parametric max-flow optimization methodology that systematically computes all breakpoints of the model in polynomial time. We also propose a data-driven class-specific prior fusion methodology, based on shape matching, alignment and fusion, that \emph{allows the shape prior to be constructed on-the-fly, for arbitrary viewpoints and partial views}. We demonstrate state of the art results in two challenging datasets: H3D\cite{PoseletsICCV09} and MPII\cite{andriluka14cvpr}, where we improve the first ranked hypothesis estimates of mid-level segmentation methods by \emph{$20\%$, with pool sizes that are up to one order of magnitude smaller.} In future work we will explore additional class-dependent seed generation mechanisms and plan to study the extension of the proposed framework to video.

\section*{Acknowledgements}
This work was supported in part by CNCS-UEFISCDI under CT-ERC-2012-1 and PCE-2011-3-0438.

{
\small
\bibliographystyle{ieee}
\bibliography{AlinPopa_ArxivPeopleSegmentation}
}

\end{document}